\documentclass[11pt]{article} 
\pdfoutput=1
\usepackage{natbib}

\usepackage[utf8]{inputenc} 
\usepackage[T1]{fontenc}    
\usepackage{hyperref}       
\usepackage{url}            
\usepackage{booktabs}       
\usepackage{amsfonts}       
\usepackage{nicefrac}       
\usepackage{microtype}      

\usepackage{lmodern}

\usepackage{graphicx}

\usepackage{amssymb}
\usepackage{amsmath}
\usepackage{amsthm}
\usepackage{caption}
\usepackage{geometry}
\newtheorem{theorem}{Theorem}

\newtheorem{defn}{Definition}
\newtheorem*{defn*}{Definition}

\newtheorem*{prop*}{Proposition}

\newtheorem{remark}{Remark}

\newenvironment{tightlist}%
{\begin{list}{$\bullet$}{%
    \setlength{\topsep}{0in}
    \setlength{\partopsep}{0in}
    \setlength{\itemsep}{0in}
    \setlength{\parsep}{0in}
    \setlength{\leftmargin}{3.5em}
    \setlength{\rightmargin}{0in}
    \setlength{\itemindent}{-.1in}
}
}%
{\end{list}
}

\newcommand{\Ex}{ \mathbb{E} }

\title{Risk-Sensitive Cooperative Games for Human-Machine Systems}

\author{
Agostino Capponi \thanks{Email: ac3827@columbia.edu, Department of Industrial Engineering and Operations Research, Columbia University, New York, NY, 10027} \and
Reza Ghanadan \thanks{E-mail: reza.ghanadan@darpa.mil, DARPA, Arlington, 22203-2114, Virginia} \and
Matt Stern \thanks{Email: mns2141@columbia.edu, Department of Industrial Engineering and Operations Research, Columbia University, New York, NY, 10027}}

\begin{document}

\maketitle

\begin{abstract}
Autonomous systems can substantially enhance a human's efficiency and effectiveness in complex environments.  Machines, however, are often unable to observe the preferences of the humans that they serve. Despite the fact that the human's and machine's objectives are aligned, asymmetric information, along with heterogeneous sensitivities to risk by the human and machine, make their joint optimization process a game with strategic interactions. We propose a framework based on {\it risk-sensitive dynamic games}; the human seeks to optimize her risk-sensitive criterion according to her true preferences, while the machine seeks to adaptively learn the human's preferences and at the same time provide a good service to the human. We develop a class of performance measures for the proposed framework based on the concept of regret.  We then evaluate their dependence on the risk-sensitivity and the degree of uncertainty. We present applications of our framework to self-driving taxis, and robo-financial advising.
\end{abstract}


\section{Introduction}
Autonomous systems can substantially enhance human effectiveness in complex environments by handling routine or cognitively challenging operations. It is crucial, however, that both the human and the machine can communicate their preferences to execute their objectives accordingly. A machine can only provide a useful or reliable service if its valuation of the costs and risks associated with each action are aligned with the human that it serves. The resulting \emph{value alignment problem} is critical to the success or failure of an operation.

In this paper, we propose a framework for the analysis of human-machine interactions. Despite the fact that the objectives of the human and the machine are aligned, there are informational asymmetries. The machine is unable to observe the human's preferences, and must infer them via a dynamic learning process by observing the effect of joint human-machine actions on the system's state. Our model is designed to capture a wide variety of situations in which a human wants to delegate a task to a machine with the objective of enhancing the efficiency and effectiveness of the task's execution. The machine is designed to serve a broad audience of humans, rather than tailored to a specific category of humans. It is thus important for the machine to personalize itself to the human and self-calibrate as the human reveals information regarding her risk preferences and objectives.

Our framework is designed to capture decision making processes arising in a large class of autonomous systems, including those for logistic operations, digital assistance, financial advising, defense, robotics, and self-driving taxi systems. As illustrative examples, consider the following two practical applications of our framework. Assume that a network of self-driving taxis tracks each time a user hails a ride. As a part of the service, the human is able to select one of the several routes or destinations that match a search query; for example, to local restaurants or retail stores. The network maintaining these cars would then keep track of each selection, marking when the human chose longer or shorter routes to higher or lower rated destinations. Our framework can generate a dynamic assessment of each user's preferences towards various destinations and on the user's sensitivity towards the risks involved in travel, such as the uncertainty of the arrival time to each destination. This assessment enables the taxicab to provide better service to the human, by presenting options that are customized for the user on the next ride.

Another example applies to the growing industry of robo-finance. Consider an investment firm that develops a portfolio-bot to manage a client's investments autonomously. In each period, the bot can reallocate the clients' investments into various assets, based on information gathered by the firm regarding the expected return and risk profile of each asset. Any allocation decision can be overridden by the client herself, through manual purchases and sales. In order to provide good service, the bot needs to understand the client's preferences for risk and return. Through our framework, the bot can estimate the preferences of the client by observing her manual investment decisions. Additionally, the firm faces risk as aggressive allocations by the bot may reflect badly on the firm if the estimates of the client's preferences are incorrect. The tolerance that the firm has towards the uncertainty over the human's preferences presents a separate form of risk-sensitivity that we define explicitly in our framework.

\paragraph{Context-driven and Human-driven risk} 

The distinguishing feature of our framework is the simultaneous handling of \emph{human-driven} and \emph{context-driven} risks. The uncertainty over the human's characteristics, such as her risk preferences, goals, and objectives, presents a human-driven risk to the machine. Depending on the machine's attitude toward risk, it could, for example, operate to provide a good service to the typical human. Alternatively, it could target humans whose characteristics belong to a specific quartile. We refer to the set of characteristics which uniquely identify the human's behavior as her type. On the other hand, the unpredictable conditions or hazardous environments in which the task needs to be executed present context-driven risks to the human. The human executes actions on the basis of her risk preferences, and in doing so it reveals information about her type to the machine. Both human and machine share the cooperative goal of minimizing the human's costs. However, informational asymmetries and heterogeneous sensitivities to risk lead to strategic behavior of the agents, and make the joint minimization process of human's costs a strategic game. In the absence of informational asymmetries, the objectives of human and machine are perfectly aligned, so that the game becomes cooperative.

\paragraph{Relation to cooperative inverse reinforcement learning} 
Our framework recovers the cooperative inverse reinforcement learning (CIRL) setup recently explored by \citet{Russell}, in the special case that there is indifference with respect to both human-driven and context-driven risk. Under these circumstances, the machine aims at providing a good service to the average human's type, and the human is only concerned about minimizing her expected costs, being neutral with respect to the risk present from the context in which she operates. The CIRL setup, however, is no longer applicable in the cases where the human is sensitive to context-driven risk or the machine is sensitive to human-driven risk.

\paragraph{Risk-sensitive equilibrium and performance measures} 
For models featuring both human-driven and context-driven risk, we introduce a new equilibrium concept, {\it risk-sensitive Bayesian equilibria}, a departure from the classical Nash equilibrium concept. Furthermore, we introduce decentralized optimization techniques to reduce the problem to a related single-agent, risk-sensitive, optimization problem. We remark that risk-sensitive optimization in the context of Markov decision processes has been subject of considerable investigation recently (see Section \ref{litreview} for additional details). In addition, our study paves the way for a systematic study of human-driven risk and its implications.

We develop a numerical study to demonstrate the power of our framework. We consider the stochastic shortest path problem with context-driven risk. This formulation captures a wide range of scenarios in which the objective of the human is to reach a goal state in the least costly way, as measured by a context-risk criterion, using actions with probabilistic outcomes. We analyze measures of performance, including the regret of our solution concept against a complete-information benchmark, and trace the impact of varying degrees of uncertainty over the human's type. 


The paper proceeds as follows. Section \ref{litreview} puts our paper in perspective with existing literature. Section \ref{sec:risav} provides a brief review of the theory of risk measures. Section \ref{sec:framework} develops the framework and presents solution concepts. Section \ref{sec:plan} provides a numerical study for the self-driving taxicab application using an abstraction based on the stochastic shortest path problem. Section \ref{sec:conclusion} concludes the paper and discusses how the proposed framework opens the door to a new class of multi-agent decision making problems.


\section{Contributions and Related Work} \label{litreview}

Our dynamic learning framework draws upon tools from machine learning, economics and finance to model human-machine interactions. The proposed framework describes the cooperative decision making problem of a human and a robot, who are both sensitive towards risk. In this section, we review related work, and put it in perspective with our contributions to the development of the framework and solution concepts

In a recent work, \citet{Russell} defines a framework for human-machine interactions, based on the theory of inverse reinforcement learning (IRL). Both the machine and the human are risk-neutral agents and, as such, their framework does not capture human-driven or context-driven risk. They reduce the two agent-model to a joint optimization problem, and compare their solution concepts to existing IRL methods. In our study, we introduce a notion of risk-sensitive equilibrium to deal with the aversion to risk of both agents in the model.


One of the defining features of our framework is that both agents share the common goal of optimizing the objective of the human. \citet{Nayyar} introduce a model of decentralized stochastic control, where a team of agents work together to minimize a common objective. They show that this problem can be reduced to a POMDP by constructing a coordinator that determines strategies for the agents, based on the common information available in each period. Similar approaches have been recently employed by \citet{Anastasopoulos-LQR} \citet{Anastasopoulos}, and \citet{Anastasopoulos-infinite} to solve incomplete information games between agents with conflicting objectives. The coordinator technique is appealing because it reduces a game of multiple agents to a single-agent optimization problem. In our framework, we show that the solution of this single-agent problem corresponds to an equilibrium between the human and machine. 

Our paper is related to existing literature on risk-sensitive Markov decision processes (MDP). Recent contributions by \citet{Bauerle-utility} and \citet{Bauerle-CVaR} solve the utility maximization process and the conditional value at risk criterion for a MDP. \citet{Haskell} generalizes these studies to a wider class of risk measures using a convex analytic approach. All these studies deal with the optimization of a single agent. In contrast, our framework features strategic interactions between agents, and employs risk-sensitive optimization to solve for a new class of equilibria corresponding to the optimal pair of human-machine actions.



\section{Background on Risk Aversion} \label{sec:risav}
In the proposed framework, humans and machines are modeled as risk-averse agents. This allows us to simultaneously capture context-driven and human-driven risk. We next provide a discussion of risk aversion, and its connection to utility functions, risk measures, and uncertainty of the outcome. In particular, utility functions are contained in the class of risk measures, and therefore we will consider risk measures in the following analysis. Informally speaking, a {\it risk-averse} agent is an agent who assigns higher weight to bad states of the world and lower weight to good states of the world, relative to a {\it risk-neutral} agent.


Consider a probability space $(\Omega, \mathcal{F}, P)$, and let the space $L^{\infty}$ be the space of essentially bounded random variables.\footnote{A random variable $Z$ is essentially bounded if there exists $M \geq 0$ such that $P(|Z| > M) = 0$.} A risk measure is a mapping $\rho$: $L^{\infty} \rightarrow \Re$ from an uncertain outcome $Z$ onto the set of real numbers. Risk measures can thus account for the entire probability distribution of an uncertain outcome, whereas expected utility functions can only depend on the realization of that outcome.

A widely used class of risk measures is the class of convex risk measures, which satisfy
\begin{enumerate}
\item Monotonicity: $\rho(Z) \geq \rho(Z')$ if $Z \geq Z'$ almost surely.
\item Translation invariance: $\rho(Z+a) = \rho(Z) + a$, if $a \in \Re$.
\item Convexity: $\rho(t Z + (1-t) Z') \leq t \rho(Z) + (1-t) \rho(Z')$ for all $t \in [0,1]$.
\end{enumerate}
The monotonicity axiom states that higher risk is associated with higher loss. The translation invariance axiom states that a sure loss of $a$ simply increases the risk by $a$.
The convexity axiom states that merging positions does not create additional risk. This property captures the benefits of diversification: a joint position containing the individual positions $X$ and $Y$ results in a lower risk overall than the sum of the risks in the position $X$ plus the risk in the position $Y$ separately.


The theory of convex risk measures can be related to utilities. Concretely, let $g: \Re \rightarrow \Re$ be a convex non-decreasing continuous disutility function, i.e. the minimum represents the point of lowest disutility. If $E[g(Z)]$ is well defined for all $Z \in L^{\infty}$, then the risk measure $\rho(Z) := E[g(Z)]$ is a convex risk measure (see also \citet{Shapiro}, chapter 6.3 therein).
{An important class of risk measures considered in this paper are those of the form}
$$J(Z) = E\left[U(Z)\right] - \theta E\left[D \left(E[Z] - Z \right) \right],$$ where $U$ is a strictly decreasing and convex utility function, and $\theta > 0$ is a coefficient quantifying risk-aversion. The function $D: \Re \rightarrow \Re^+$ measures the deviation of the losses from the expectation. If we choose $U(Z) = -Z$ and $D(Z) = Z^2$, we recover the classical mean-variance problem. If we set $D$ to be a convex function with $D(Z) = 0$ for $Z \geq 0$, then the second component of $J(Z)$ captures downside risk of losses.



\section{The Framework} \label{sec:framework}

\begin{defn}

A human-machine interaction game is a $T$ period, dynamic game with asymmetric information played between two risk sensitive agents: a human, {\bf H}, and a machine, {\bf M}. The game is described by a tuple $\left\langle \mathcal{S}, \left\{\mathcal{A}^{\mathbf H},\mathcal{A}^{\mathbf M}\right\}, \Theta, \left\{ \rho^{\mathbf{H}}, \rho^{\mathbf{M}}\right\}, P, c, \pi_{0}\right\rangle$, whose elements are defined as:

\begin{tightlist}
\item[$\mathcal{S}$] a set of system states: $s \in \mathcal{S}$;
\item[$\mathcal{A}^{\mathbf{H}}$] a set of actions for $\mathbf{H}$: $a^{\mathbf{H}} \in \mathcal{A}^{\mathbf{H}}$;
\item[$\mathcal{A}^{\mathbf{M}}$] a set of actions for $\mathbf{M}$: $a^{\mathbf{M}} \in \mathcal{A}^{\mathbf{M}}$;
\item[$\Theta$] a set of possible risk parameters, only observed by $\mathbf{H}$: $\theta \in \Theta$;
\item[$\rho_{\theta}^{\mathbf{H}}(\cdot)$] $\mathbf{H}$'s convex risk measure, parameterized by $\theta$;
\item[$\rho^{\mathbf{M}}(\cdot)$] $\mathbf{M}$'s convex risk measure over a probability distribution on $\theta$;
\item[$P(\cdot | \cdot, \cdot, \cdot)$] the probability transition function on the future state, given the current state and joint action: $P(s' | s, a^{\mathbf{H}}, a^{\mathbf{M}})$;
\item[$c(\cdot, \cdot, \cdot)$] an instantaneous cost function that maps the system state and joint actions to a vector of real numbers: $c:\mathcal{S}\times \mathcal{A}^{\mathbf H} \times \mathcal{A}^{\mathbf R} \rightarrow \Re^{N}$;
\item[$\pi_{1}(\cdot) $] a common prior distribution over the risk parameters: $\pi_{1}(\theta) \in  \mathcal{P}(\Theta)$.
\end{tightlist}
\end{defn}
We allow for $N$ different cost drivers, such as time, labor, or consumed materials depending on the application considered. In a self-driving taxis application, the system costs may include the time required to drive the passenger to her destination, and the toll amount paid by the passenger to the taxi driver. For example, a high toll bridge would allow the passenger to arrive faster to her destination, but it will require him to pay additional fees.
Above, we have used $\mathcal{P}(\Theta)$ to denote the set of probability distributions on $\Theta$. After each period $t$, the human and the machine incur $N$ common costs, $c(s_{t},a^{\mathbf{H}}_{t},a^{\mathbf{M}}_{t})\in \Re^{N}$, depending on the current state of the system, and their joint action. Their incentives are partially aligned as both the human and the machine prefer to keep the total system costs low over the $T$ period horizon. The human's objective is to minimize the $N$ costs using her risk measure as the optimization criterion $\rho^{\mathbf{H}}_{\theta}$, where $\theta$ is the true type of the human. For example, the mean-variance risk measure $\rho_{\theta}^{\mathbf{H}} =
\theta_{1}^{\top}\Ex\left[\sum_{\tau=1}^{T}c(s_{\tau},a^H_\tau,a^M_\tau) \right] + \theta_{2}^\top Var\left[\sum_{t=1}^{T}c(s_{\tau},a^H_\tau,a^M_\tau)\right]$ maps an $N$ dimensional random outcome for the total costs to a scalar quantity through the parameter vector $\theta = [\theta_{1}, \theta_{2}] \in \Re^{2N}$. In this case, the vector $\theta$ not only describes the human's risk sensitivity towards each of the $N$ costs, but it also describes the relative weight that the human assigns to each cost. The machine does not know the value of $\theta$ at the initial stage of the game, but begins with a prior distribution $\pi_1(\cdot)\in \mathcal{P}(\Theta)$. The machine's objective is to minimize the risk measure criterion $\rho^{\mathbf{M}}$.

Denote the set of public histories as $$H_{t} := \left(\mathcal{A}^{\mathbf{H}}\times\mathcal{A}^{\mathbf{M}}\right)^{t-1}\times\mathcal{S}^{t},$$ where $h_{t} = \left(s_{1},a^{\mathbf{H}}_{1}, a^{\mathbf{M}}_{1}, \ldots, a^{\mathbf{H}}_{t-1}, a^{\mathbf{M}}_{t-1}, s_{t}\right) \in H_{t}$ for $t> 1$ and $h_1 = s_1$. A public history contains information that is observed by both the human and the machine, which includes the realization of the system's states and the actions executed by both agents. The machine maintains the posterior distribution over the human's type, $\pi_t(x) := P(\theta = x |h_t)$, which we refer to as the machine's belief in period $t$.

A Markov strategy for the human $\sigma^{{\mathbf{H}}} = (\sigma^{{\mathbf{H}}}_{1}, \ldots, \sigma^{{\mathbf{H}}}_{T})$  is a sequence of measurable maps $\sigma_{t}^{{\mathbf{H}}} : \mathcal{S} \times \mathcal{P}(\Theta) \times \Theta \rightarrow \mathcal{P}\left(\mathcal{A}^{{\mathbf{H}}}\right)$ so that
\begin{eqnarray*}
&\sigma_{t}^{\mathbf{H}}(a|s_{t},\pi_t,\theta) = P(a_{t}^{\mathbf{H}}=a|s_{t}, \pi_t, \theta), \; \; \forall t\in \{1,\ldots,T\}, a \in \mathcal{A}^{\mathbf{H}}.
\end{eqnarray*}
A Markov strategy for the machine  $\sigma^{{\mathbf{M}}} = (\sigma^{{\mathbf{M}}}_{1}, \ldots, \sigma^{{\mathbf{M}}}_{T})$  is a sequence of measurable maps $\sigma_{t}^{{\mathbf{M}}} : \mathcal{S} \times \mathcal{P}(\Theta)  \rightarrow \mathcal{P}\left(\mathcal{A}^{{\mathbf{M}}}\right)$ so that
\begin{eqnarray*}
\sigma_{t}^{\mathbf{M}}(b|s_{t}, \pi_t) = P(a_{t}^{\mathbf{M}}=b |s_{t}, \pi_t), \; \; \forall t\in \{1,\ldots,T\}, b \in \mathcal{A}^{\mathbf{M}}
\end{eqnarray*}
Notice that the human's Markov strategy depends on the machine's current beliefs because the action of the human is influenced by the action of the machine, which in turn depends
on its belief over the human's type.

The total (cumulative) cost is given by the random variable $$C_{T} := \sum_{\tau = 1}^{T} c\left(s_{\tau}, a_{\tau}^{\mathbf{H}}, a_{\tau}^{\mathbf{M}}\right).$$

Given the conflicting objectives of both agents, the framework as presently defined is a two-player strategic game. As such, we define the corresponding {\it risk-sensitive Bayesian equilibrium} as a pair of strategies $\left(\sigma^{*\mathbf{H}}, \sigma^{*\mathbf{M}}\right)$ and a belief profile $\pi^{*} := (\pi^{*}_1, \ldots,\pi^{*}_T)$ such that
\begin{eqnarray}
\nonumber \rho_{\theta}^{\mathbf{H}}\left(C_{T} | \sigma^{\mathbf{*H}},\sigma^{\mathbf{*M}}, \pi_1^*, h_1 \right) &\leq& \rho_{\theta}^{\mathbf{H}}\left(C_{T} |{\tilde\sigma^{\mathbf{H}},\sigma^{\mathbf{*M}}, \pi_1^*, h_1}\right), \\
\rho^{\mathbf{M}}\left(\rho^{\mathbf{H}}_{\theta}\left(C_{T} | {\sigma^{\mathbf{*H}},\sigma^{\mathbf{*M}}}, \pi_1^*, h_1 \right)  | \pi_1^* \right) &\leq& \rho^{\mathbf{M}}\left(\rho_{\theta}^{\mathbf{H}}\left(C_{T} | {\sigma^{\mathbf{*H}},\tilde\sigma^{\mathbf{M}}}, \pi_1^*,h_1 \right) | \pi_1^*  \right),
\label{eq:rhoHM}
\end{eqnarray}
for all strategies $\tilde\sigma^{\mathbf{H}}$, $\tilde\sigma^{\mathbf{M}}$. Furthermore, the machine's belief profile $\pi^{*}$ must be consistent with the strategies $\left(\sigma^{*\mathbf{H}}, \sigma^{*\mathbf{M}}\right)$ in that Bayes' rule is used to update the beliefs whenever possible. Specifically, the machine's belief on the true value of the human's risk parameter $\theta$ satisfy the standard nonlinear filter equation (\citet{FudenbergTirole}),
\begin{eqnarray}
\pi^{*}_{t+1}(\theta) := \dfrac{\pi^{*}_{t}(\theta )\sigma^{*\mathbf{H}}(a_{t}^{\mathbf{H}}|s_{t}, \pi^{*}_t, \theta)}{\sum_{\tilde\theta}\pi^{*}_{t}(\tilde\theta)\sigma^{*\mathbf{H}}(a_{t}^{\mathbf{H}}|s_{t}, \pi^{*}_t, \tilde\theta)}, \label{eq:belief}
\end{eqnarray}
provided there exists a value of $\tilde{\theta}$ such that $\pi^{*}_{t}(\tilde{\theta}) > 0$ and $\sigma^{*\mathbf{H}}(a_{t}^{\mathbf{H}}|s_{t}, \pi^{*}_t, \theta) > 0$. In period 1, the belief profile $\pi^*_1$ is equal to the prior $\pi_1$.

The first of the two inequalities in equation \eqref{eq:rhoHM} indicates that the human has no incentive to unilaterally deviate from her action $\sigma^{\mathbf{*H}}$ to any other action $\tilde\sigma^{\mathbf{H}}$ because her risk-adjusted total cost would increase. Similarly, the second inequality stipulates that the machine's action yields the smallest risk-adjusted total cost, according to both the human's type and the machine's beliefs over the human's type.

The canonical solution concept for dynamic games of incomplete information is the Bayesian equilibrium (BE). However, standard equilibrium concepts rely on maximizing the expectation of utility functions assigned to each player.  A Bayesian equilibrium in our setup would require that both agents minimize the expected disutility of total system costs, rather than the general risk measures we present.

Context-driven risk in our model is captured by applying the risk measure $\rho^{\mathbf{H}}$ to the total system cost. This allows us to capture a wide variety of cost criteria that depend on the statistical properties of the cumulative costs, including value at risk, conditional value at risk, and worst case measures. A special case of context-driven risk is when the human minimizes the expected disutility from costs, where disutility is quantified by a convex utility function. Human-driven risk is quantified by the risk measure $\rho^{\mathbf{M}}$ over the distribution of human's type. For example, if $\rho^{\mathbf{M}}$ is the expectation operator, then the machine aims for the best service to the average human type. On the other hand, if $\rho^{\mathbf{M}}$ represents the value at risk for some level of service $\alpha$, then the machine aims to provide good service for $1-\alpha$ percentage of the human types. Lastly, if the human's type is revealed to the machine before $T$, then there is no human's driven risk. In this case, $\rho^{\mathbf{M}}\left(\rho^{\mathbf{H}}_{\theta}\left(C_{T} \right) \right) = \rho^{\mathbf{H}}_{\theta}\left(C_{T} \right)$, so that the two inequalities in Eq.~\ref{eq:rhoHM} coincide, and the game becomes {\it cooperative}.

The solution methodology that we propose to address the human-machine framework is to transform the strategic game to a single-agent problem by introducing a coordinator agent $\mathbf{C}$. The coordinator assigns a policy $\sigma^{\mathbf{C}} = (g^{\mathbf{M}}, g_{\theta}^{\mathbf{H}})$ such that $g^{\mathbf{M}}$ is a strategy for the machine and $g_{\theta}^{\mathbf{H}}$ is a decision function, which prescribes the human's strategy for each possible realization of $\theta$. Hence, the coordinator is unaware of the human's risk parameter, but instead chooses a strategy for every possible type of human. The coordinator's objective is to minimize the machine's risk measure using these controls,
\begin{eqnarray*}
\min_{g^{\mathbf{M}}, g_{\theta}^{\mathbf{H}}} \rho^{\mathbf{M}}\left(\rho_{\theta}^{\mathbf{H}}\left(C_{T}, |  g_{\theta}^{\mathbf{H}}, g^{\mathbf{M}},\pi_1,h_1 \right) | \pi_1  \right).
\end{eqnarray*}
The resulting problem is a partially-observable, risk-sensitive, Markov decision process (risk-POMDP). The following theorem connects the solution to the coordinator problem with the equilibrium concept for the human-machine interaction game.

\begin{theorem}\label{thm:coordinator}
A solution to the coordinator problem is a risk-sensitive Bayesian equilibrium to the two-agent human-machine interaction game.
\end{theorem}

The proof of Theorem \ref{thm:coordinator} is included in appendix \ref{sec:proof}.

\begin{remark}
The risk-POMDP formulation can be reduced to a fully observable, risk-sensitive MDP, where the state space is enlarged to include the belief profile. The resulting single-agent problem can then be solved using existing risk optimization techniques described in Section \ref{litreview}.
\end{remark}


\section{Applications and Numerical Study} \label{sec:plan}


Consider the following risk-sensitive version of the stochastic path finding problem. A human hires a self-driving taxicab to travel to one of several possible locations, but each path assumes an uncertain cost for the human. This cost may take the form of a fuel consumption cost required to travel, tolls or fees required for access to certain paths, or a cost associated with the time the car takes to reach one of the destinations. In each period, the car chooses a direction to move along a grid towards a destination. Making this determination requires the car to estimate the human's risk preferences associated with the cost of travel. For instance, if the human's tolerance for risk was known to be high, then the car would choose a path with a lower expected cost, irrespective of its risk level. Conversely, if the robot knew that the human was sensitive to risk, it would avoid paths with widely varying costs, even if those paths had lower average costs.

As an illustrative example, assume that the human's risk aversion coefficient is $\theta$ and that the optimal path minimizes the weighted sum of its expected cost and variance. Note that, for this example, the risk-aversion coefficient is the human's type.
\begin{equation}
\Ex\left[\sum_{\tau=1}^{T}c(s_{\tau},a^H_\tau,a^M_\tau) \right] + \theta Var\left[\sum_{t=1}^{T}c(s_{\tau},a^H_\tau,a^M_\tau)\right],
\label{eq:riskaverseopt}
\end{equation}
where each $c(s_{\tau},a^H_\tau,a^M_\tau)$ represents the random cost of traveling from gridpoint $s_{\tau}$ to the gridpoint determined by the joint actions {$a^H_\tau$ and $a^M_\tau$}. Once the car stops moving, the human receives a reward (or negative cost) associated with its final position $c(s_{\tau},STOP) < 0$. Assuming that $\theta$ is known, the optimal solution solves for the \emph{stochastic shortest path} adjusted for risk aversion. The figure below depicts optimal paths for two values of the risk aversion parameter, with $\theta_{1} < \theta_{2}$ representing a stronger aversion to the variance of total travel costs. Each edge $(i,j)$ is labeled with the mean and variance of its cost $(\mu_{ij}, \sigma_{ij}^{2})$.

\begin{figure}[ht!]
\begin{center}
 \includegraphics[width=5.5cm]{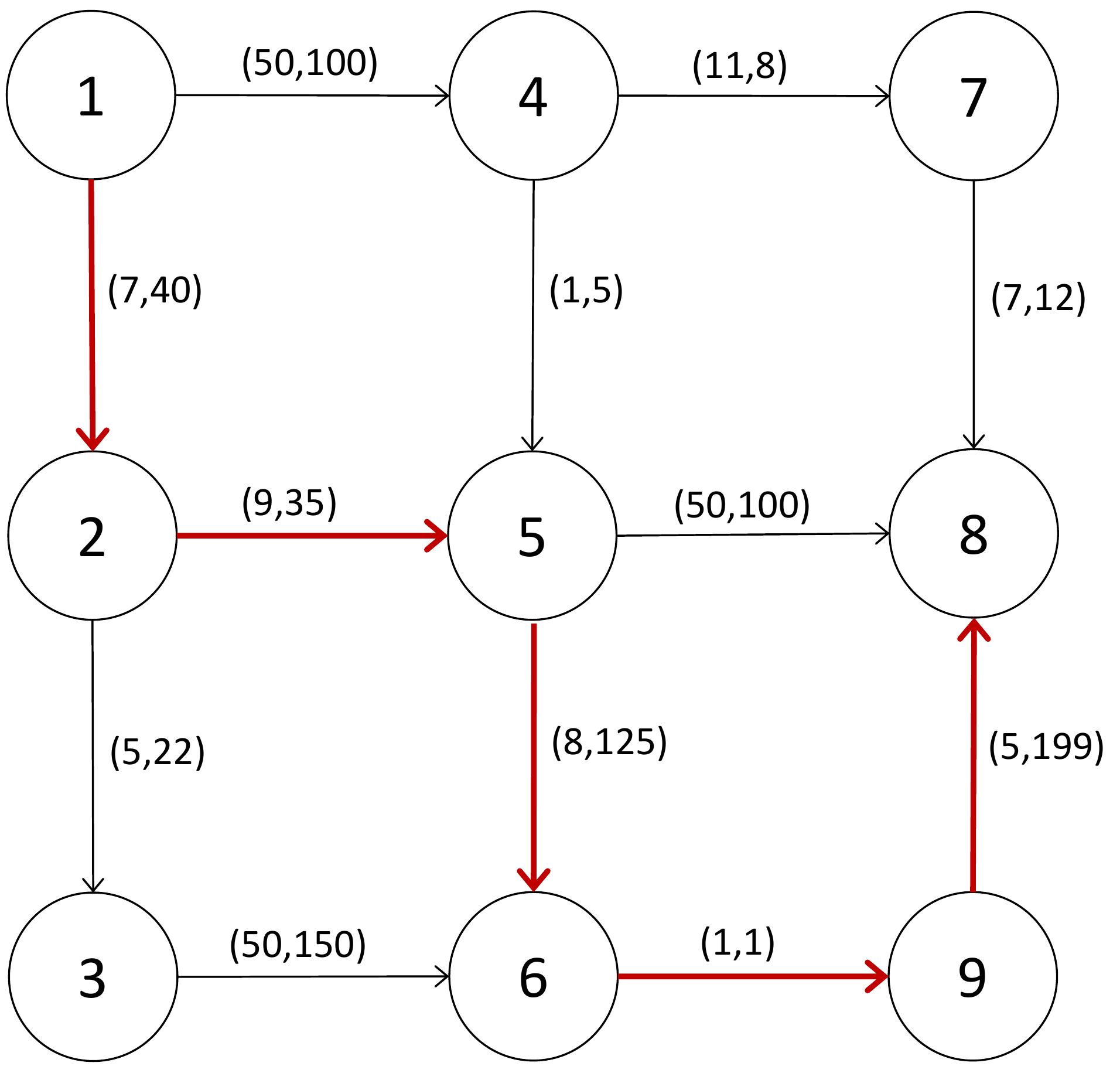} \hspace{2cm}
 \includegraphics[width=5.5cm]{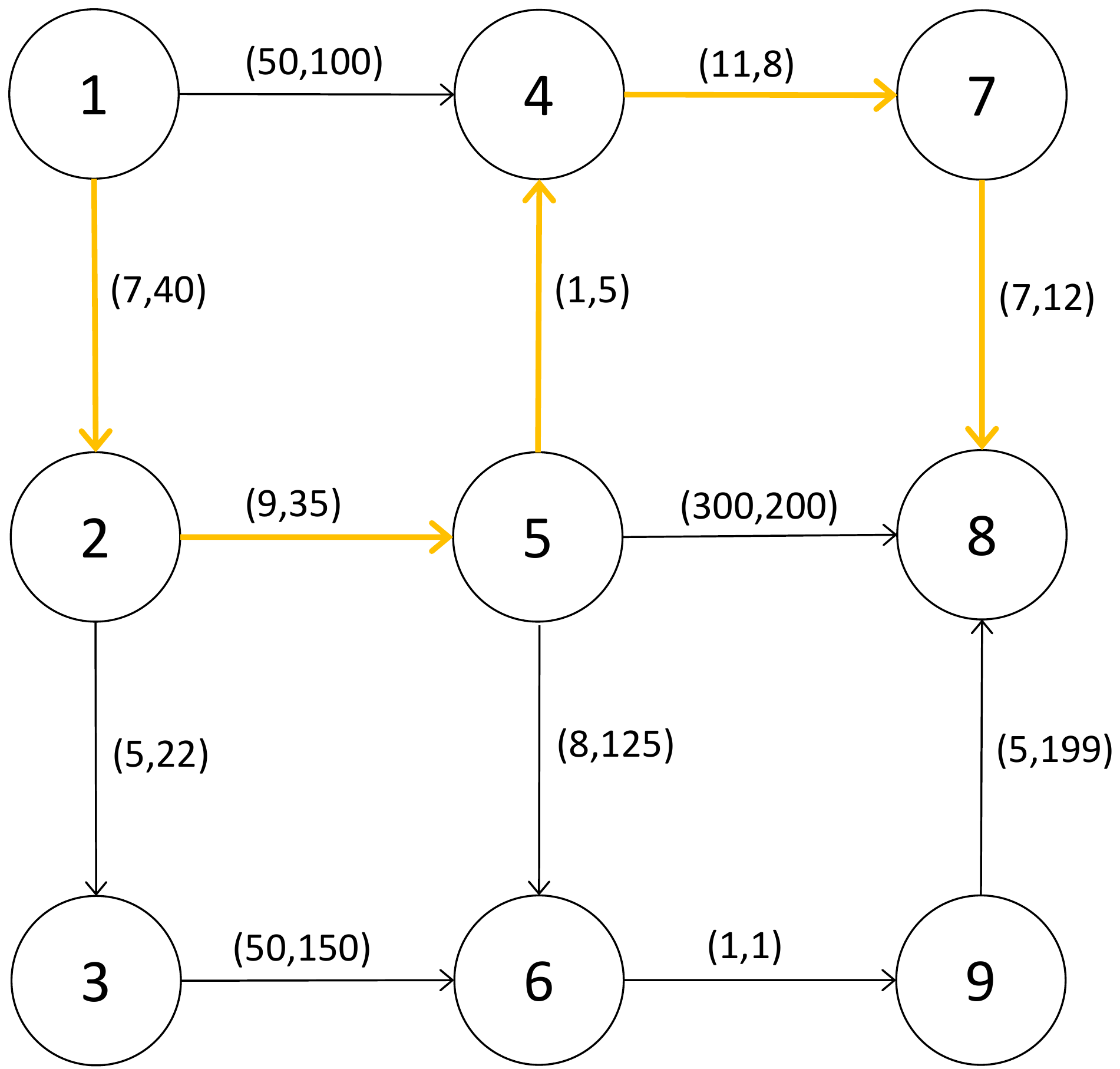}\\
\end{center}
\caption{{\it The human's planning problem.} The costs are assumed to be independent across edges. The objective is to minimize the risk-averse criterion in Eq.~\eqref{eq:riskaverseopt}, that is to solve the stochastic shortest path from node 1 to node 8.\\
Left: the shortest path if the human's risk aversion coefficient is $\theta_1 = 0.01$ (red arrows denote the shortest path). The risk averse criterion equals $30 + 0.01 \times 400 = 34$. Right: the shortest path if the human's risk aversion coefficient is $\theta_2 = 0.05$ (yellow arrows denote the shortest path). The risk averse criterion equals $35 + 0.05 \times 100 = 40$.}
\label{figstochshortpath}
\end{figure}

Each of these paths would be followed by the car, provided it knew the human's risk aversion before proceeding along a chosen direction. Instead the machine is calibrated, perhaps by its manufacturer, with a prior distribution over the two possible risk coefficients, {$\pi_{1}(\theta_{1}) = Pr(\theta = \theta_{1})$}, {and $\pi_{1}(\theta_{2}) = Pr(\theta = \theta_{2})$}. Left unattended by the human, the car would proceed along the path minimizing the risk-adjusted cost corresponding to the expected risk-aversion coefficient, {$\pi_{1}(\theta_{1})\theta_{1} + \pi_{1}(\theta_{2})\theta_{2}$}. Accordingly, the car chooses an action $a^M(t) \in \left \{ ``N'', ``S'', ``E'', ``W'', ``STOP'' \right \}$ in each period. The human observes the car as it moves and decides in each period whether or not to send a signal to override the car's action and replace it with one that better reflects the human's true risk aversion parameter. The human's action in period $t$, $a^{H}(t) \in \left \{\emptyset, ``N'', ``S'', ``E'', ``W'', ``STOP'' \right \}$ incurs a fixed transmission cost $q(a^H(t)) = q^{H}\left({\bf 1}_{a^H(t) \neq \emptyset}\right)$ each time a signal is sent to the car.

Whenever the human overrides the car's action, the car receives more information about the human's underlying risk parameter. Consider once again, the example illustrated in the figure. Before period $t=3$, the risk-adjusted shortest path is the same for both $\theta_{1}$ and $\theta_{2}$. The human would therefore not send a signal to the car in these periods, since the car would execute the optimal path without additional and costly guidance. In period $t=3$, however, the human would direct the car towards the optimal path with $a^{H}(3) = ``S''$ if $\theta = \theta_{1}$ and $a^{H}(3) = ``N''$ if $\theta = \theta_{2}$. After the human sends this action to the car, the car would learn the human's true risk parameter as this action is fully revealing. The human would then refrain from sending any more transmissions, as the car would act optimally according to the human's revealed preferences until reaching the destination.

\begin{figure}[ht!]
\begin{center}
\includegraphics[width=5.5cm]{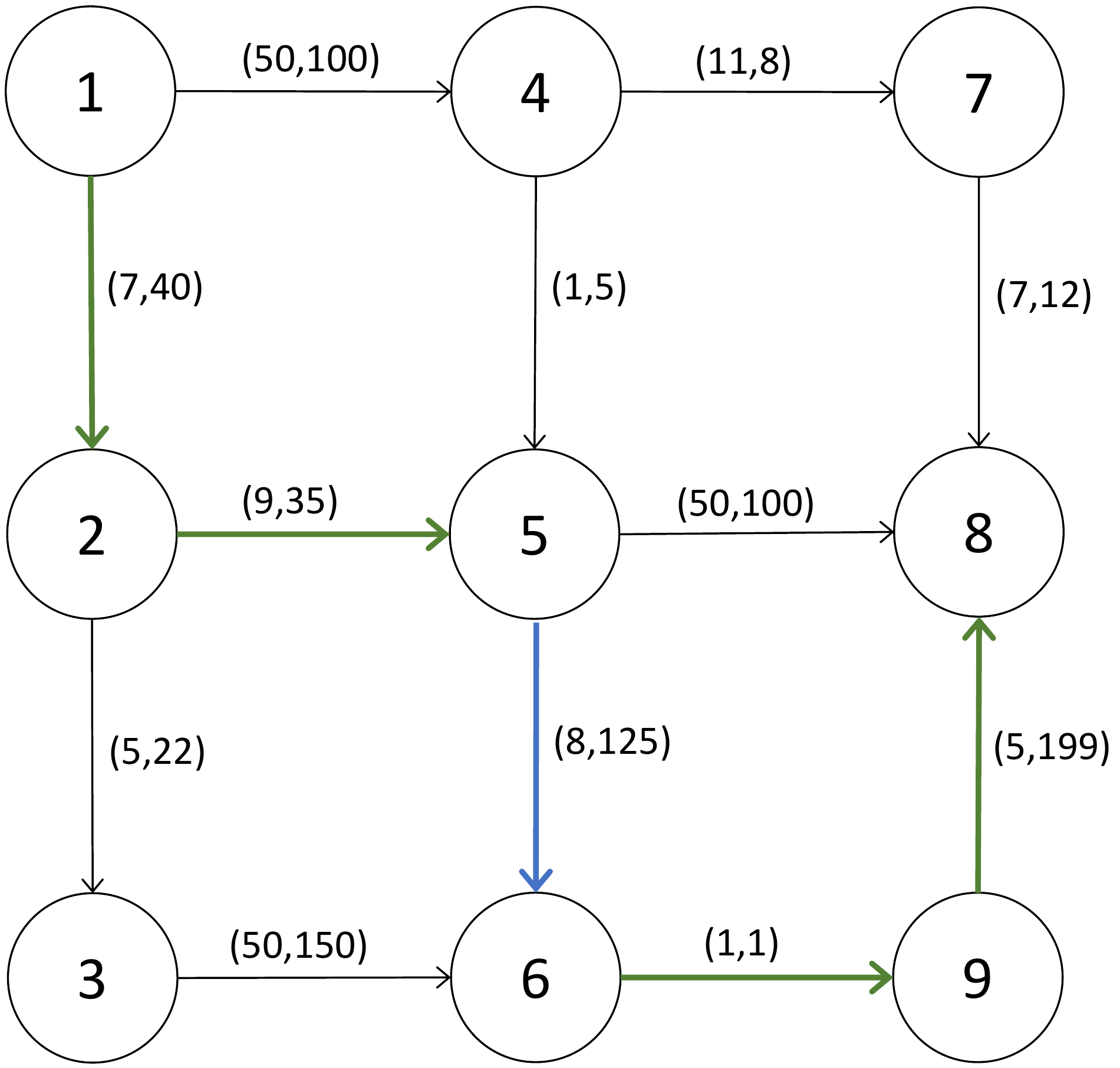} \hspace{2cm}
\includegraphics[width=5.5cm]{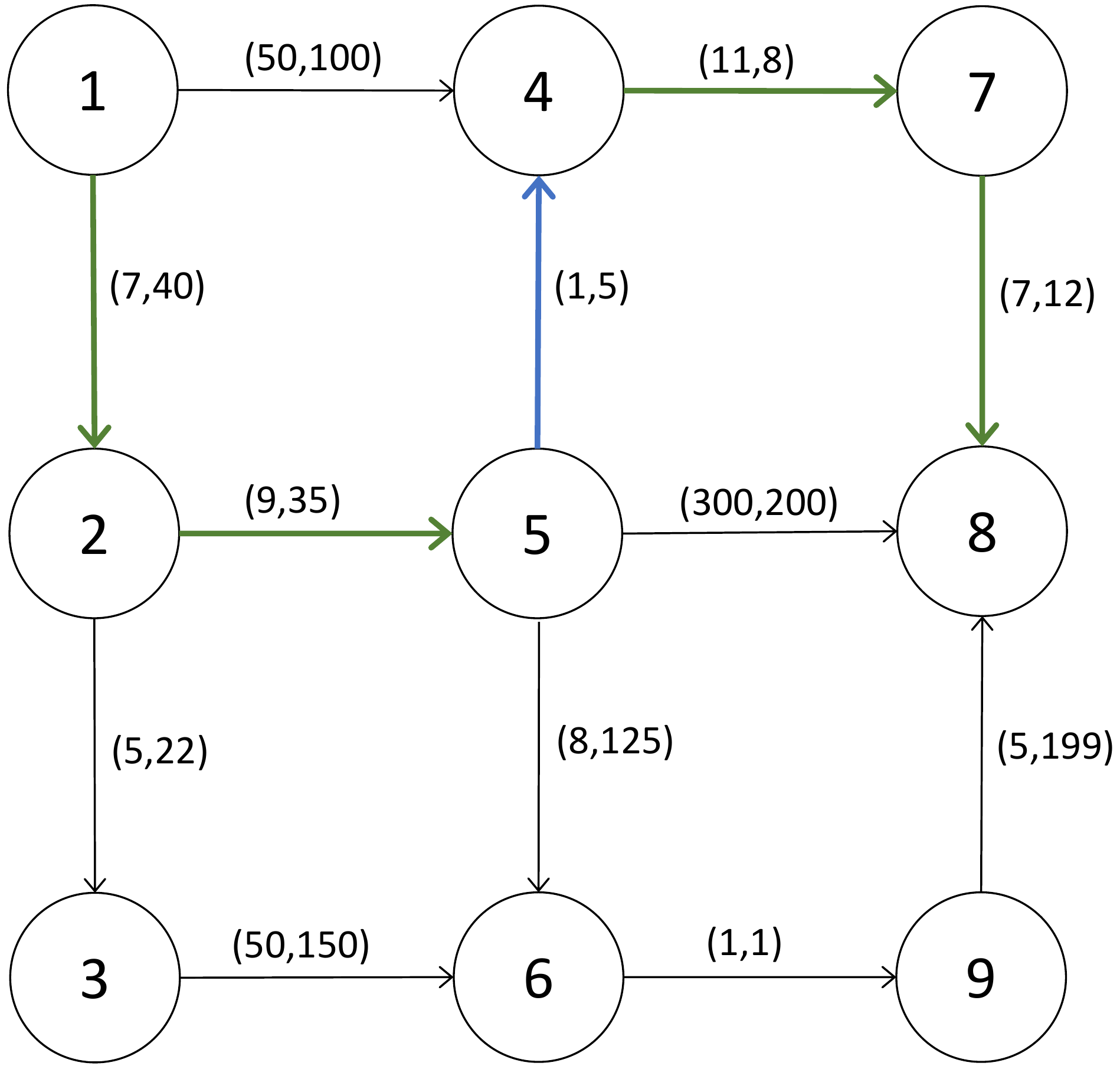}
\end{center}
\caption{{\it The human-machine planning problem.} Human and machine cooperate to minimize the risk-averse criterion in Eq.~\eqref{eq:riskaverseopt}, that is the stochastic shortest path from node 1 to node 8. Green arrows correspond to machines's actions and blue arrows correspond to human's actions. \\
Left: The actual human's risk aversion coefficient is $\theta_1 = .01$. Right: The actual human's risk aversion coefficient is $\theta_2 = .05$.\\ The human's action is fully revealing and allows the machine to perfectly learn the human's risk aversion coefficient.}
\label{figstochshortpath}
\end{figure}

\subsection{Measures of Performance}

\begin{figure}[ht!]
\includegraphics[width=7.2cm]{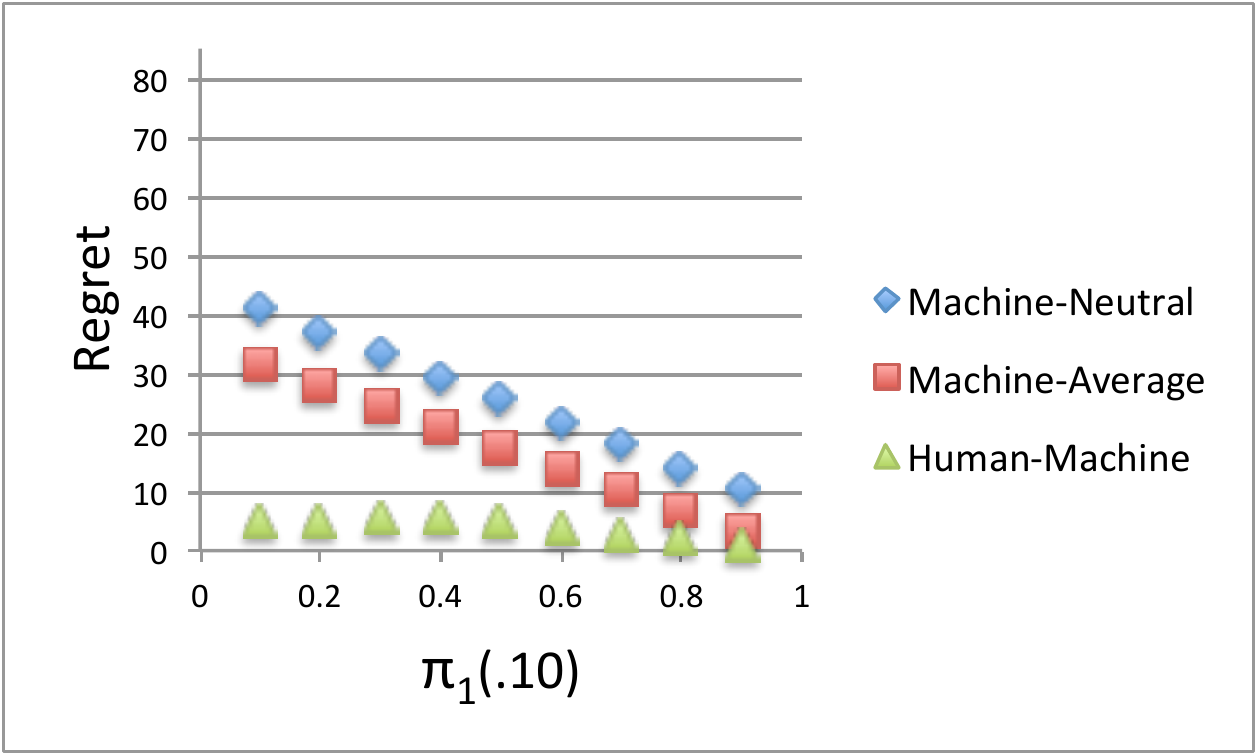}
\includegraphics[width=7.2cm]{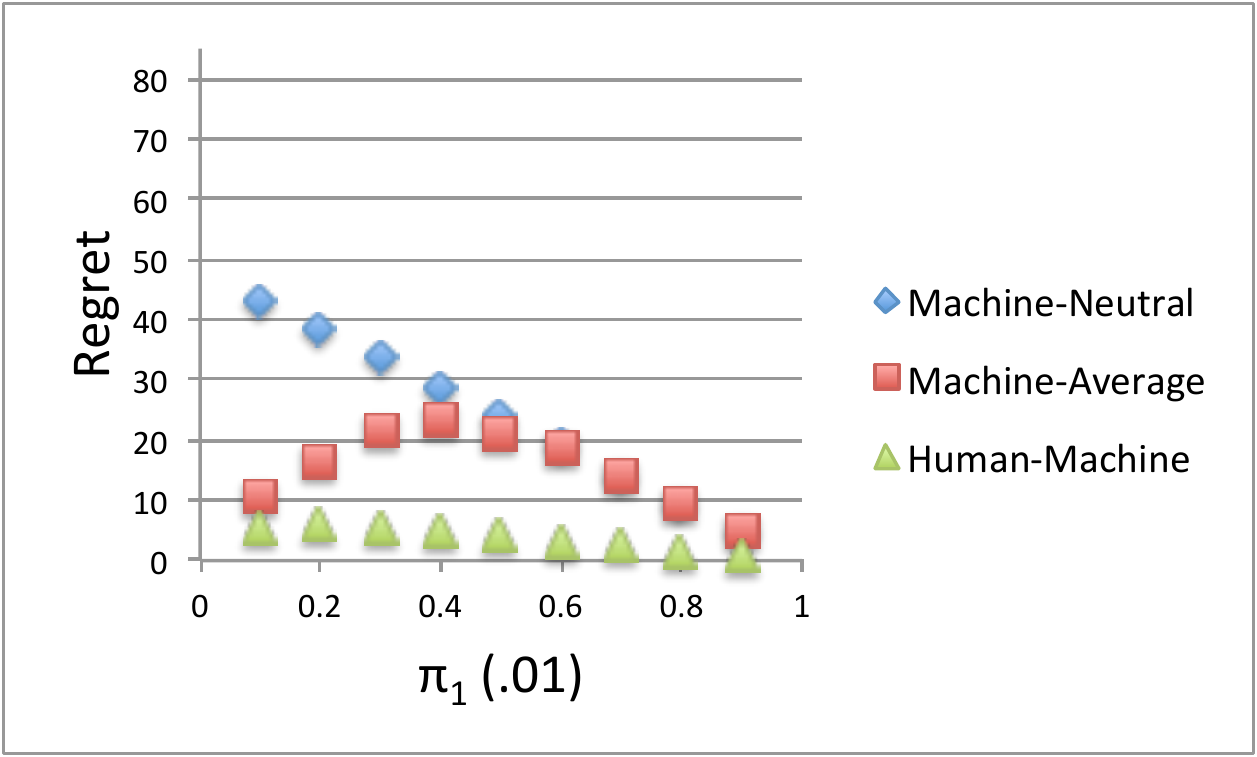}
\begin{center}
\includegraphics[width=7.2cm]{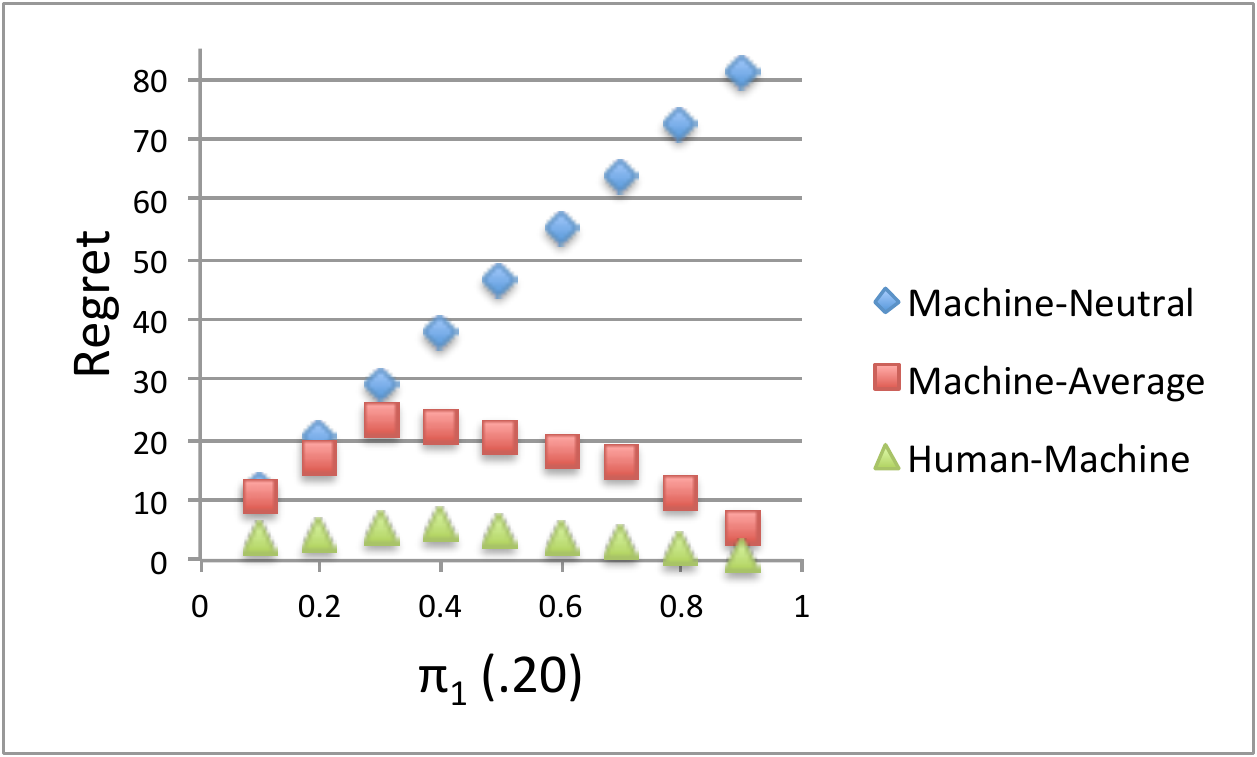}
\end{center}
\caption{\it Each graph plots the regret of the different strategies versus the prior probability on the actual human's risk aversion parameter. Top left graph: vary $\pi_1(\theta_1)$ on the x-axis and assume $\pi_1(\theta_2)=\pi_1(\theta_3) = \frac{1-\pi_1(\theta_1)}{2}$. Top right graph: vary $\pi_1(\theta_2)$ on the x-axis and assume $\pi_1(\theta_1)=\pi_1(\theta_3) = \frac{1-\pi_1(\theta_2)}{2}$. Bottom graph: vary $\pi_1(\theta_3)$ on the x-axis and assume $\pi_1(\theta_1)=\pi_1(\theta_2) = \frac{1-\pi_1(\theta_3)}{2}$.}
\label{figstochshortpath}
\end{figure}
To measure the effectiveness of the coordinator risk-POMDP, we assess the following measures of performance. Using the self-driving taxicab framework, consider the case in which the car is undecided between three possible values of human's risk aversion: $\theta_1 = 0.01, \theta_2 = 0.1$, and $\theta_3 = 0.2$. The mean and variance of the edges can be found in appendix \ref{sec:numbericaldetails}. We calculate the \emph{regret} of various strategies, including the risk-POMDP solution, against the best possible outcome. This benchmark is defined to be the solution in a setup where the human is allowed to directly communicate her risk parameter to the machine in the first period. The ``Machine-Neutral'' strategy represents the policy of a car, which ignores the risk of each path; that is, the car assumes incorrectly that $\theta = 0$. The ``Machine-Average'' strategy uses the correct prior over the human's risk parameter, but does not interact with the human while it drives. As such, the human never gets the opportunity to reveal more information about her risk parameter, and the car provides the best service to the average human. This policy is similar to the strategies considered in \citet{Russell} for apprenticeship learning models. In these models, the machine observes the human for a fixed number of periods and generates a probability distribution over the human's type. The machine then takes actions to optimize the objective of the average human according to the prior distribution on the human's type. The machine-average strategy mirrors this learning model in the sense that the machine's prior could be generated from earlier interactions with the human. Lastly, the ``Human-Machine'' strategy solves the risk-POMDP in a setup where the human is allowed to override the machine's actions and teach the car her risk preferences. Mathematical formulations for each policy are available in appendix \ref{sec:numbericaldetails}.

\subsection{Numerical Results}
The graphs in Figure~\ref{figstochshortpath} depict the regret from the different strategies. The optimal joint human-machine policy always achieves the smallest regret. Moreover, the regret from the strategy in which the machine ignores the context-driven risk is the highest. Consistent with the intuition, the left graph indicates that if the prior probability on the lowest value of human's risk aversion increases, the regret from not accounting for the risk parameter in the machine's strategy gets lower. Consequently, the machine-neutral strategy has a small regret because the true sensitivity to context-driven risk is very low ($\theta=0.01 \approx 0$). In contrast, if the prior probability on the highest value of human's risk aversion increases (right graph), the regret of the strategy in which the machine does not account for risk becomes higher. In this case, ignoring risk is costly because the human is very sensitive to context-driven risk ($\theta=0.2$).

The machine-average strategy outperforms the machine-neutral strategy in each instance since it incorporates the typical (average) human's preference towards context-driven risk. As the prior probability on the human's risk aversion parameter approaches a point mass distribution, i.e. $\pi_1(\theta)$ increases toward one for a fixed $\theta$, under the machine-average strategy, the machine provides the same service as in the best case policy, in which it knows the human's risk aversion. When the prior probability for one of the parameters is low, the machine-average strategy takes the average of the remaining two parameter values to determine its policy. As a result, the regret when $\pi_1(.01) \approx 0$ shown in the left graph is lower than the regret for intermediate values of the prior. Low values of $\pi_1(.01)$ imply that the true parameter is either $\theta = .1$ or $\theta = .2$ and so the optimal policy corresponding to $\bar \theta = .15$ performs closer to the best case policy than, for example, the optimal policy when $\pi_1(.01) = .33$, which corresponds to $\bar \theta = .1033$.

\section{Conclusion and Future Work} \label{sec:conclusion}
In this paper, we presented a framework for human-machine decision making, accounting for both human-driven and context-driven risk. Due to the different risk sensitivities of the human and the machine, respectively, to the context in which the task is being executed and to the category of humans served, the optimal decision making problem may be formulated as a game with strategic interactions. We have introduced the concept of risk-sensitive equilibria to deal with the corresponding game, and introduced risk-sensitive optimization techniques to solve a related coordinator problem. The optimal solution to the risk-POMDP is then a risk-sensitive Bayesian equilibrium for the human-machine framework. Using our framework, we have developed measures of performances quantifying the regret of various human-machine cooperative strategies against the idealized situation in which both human and machine know human's characteristics. Our numerical analysis indicates that human-machine strategies may exhibit poor performance if they ignore context-driven risk when this is actually present, or if they target the profile of human having the typical (average) characteristics.


Future directions for this research include the development of new solution methods to integrate risk optimization techniques with concepts from game theory. A key refinement to equilibrium in dynamic games is the notion of subgame perfection. This enforces incentive compatibility for both agents in each subgame initiated at the start of each period. However, many commonly used risk measures are not time-separable, i.e. the risk over the entire horizon cannot be decomposed into a set of risks, each allocated to a different time period. Consider, for example, a self-driving taxi application, in which heavy traffic conditions at a specific hour of the day are likely to be followed by severe traffic conditions at nearby time periods. In this situation, risks are correlated over time, and thus an error would be introduced if they were to be treated as time separable. Without time separability, the risk-POMDP no longer satisfies the Markov property. 
\appendix
\section*{Appendix} \label{sec:appendix}

\section{Proof of Theorem 1} \label{sec:proof}
Given a solution to the coordinator problem, $\sigma^{\mathbf{*C}} = (g^{*\mathbf{M}},g_{\theta}^{*\mathbf{H}})$, assume that the human's true type is an arbitrary value $\bar\theta \in \Theta$. Then it is sufficient to confirm that the strategies $\sigma^{\mathbf{*H}} = g_{\bar\theta}^{\mathbf{*H}}$ and $\sigma^{*\mathbf{M}}=g^{\mathbf{*M}}$ satisfy the three properties for the risk-sensitive Bayesian equilibrium:
\begin{itemize}
\item[(I)] Machine's incentive compatibility
$$\rho^{\mathbf{M}}\left(\rho^{\mathbf{H}}_{\theta}\left(C_{T} | {\sigma^{\mathbf{*H}},\sigma^{\mathbf{*M}}}, \pi_1, h_1 \right)  | \pi_1 \right) \leq \rho^{\mathbf{M}}\left(\rho_{\theta}^{\mathbf{H}}\left(C_{T} | {\sigma^{\mathbf{*H}},\tilde\sigma^{\mathbf{M}}}, \pi_1,h_1 \right) | \pi_1  \right).$$
\item[(II)] Human's incentive compatibility
$$\rho_{\bar\theta}^{\mathbf{H}}\left(C_{T} | \sigma^{\mathbf{*H}},\sigma^{\mathbf{*M}}, \pi_1, h_1 \right) \leq \rho_{\bar\theta}^{\mathbf{H}}\left(C_{T} |{\tilde\sigma^{\mathbf{H}},\sigma^{\mathbf{*M}}, \pi_1, h_1}\right).$$
\item[(III)] The consistent belief profile
\begin{eqnarray}
\pi^{*}_{t+1}(\theta) := \dfrac{\pi^{*}_{t}(\theta )\sigma^{*\mathbf{H}}(a_{t}^{\mathbf{H}}|s_{t}, \pi^{*}_t, \theta)}{\sum_{\tilde\theta}\pi^{*}_{t}(\tilde\theta)\sigma^{*\mathbf{H}}(a_{t}^{\mathbf{H}}|s_{t}, \pi^{*}_t, \tilde\theta)}. \label{eq:belief}
\end{eqnarray}
\end{itemize}

The machine's incentive compatibility (I) is satisfied since the objective function of the coordinator is equal to the objective function of the machine. The consistent belief profile (III) follows directly from the formulation of the coordinator problem as a POMDP. The human's incentive compatibility condition (II) follows from the monotonicity property of the convex risk measures $\rho_{\theta}^{\mathbf{H}}$ and $\rho^{\mathbf{M}}$ by the following logic. Assume that there exists a strategy $\tilde{\sigma}_{\bar\theta}^{\mathbf{H}}$ such that $$\rho_{\bar\theta}^{\mathbf{H}}\left(C_{T} | \sigma^{\mathbf{*H}},\sigma^{\mathbf{*M}}, \pi_1, h_1 \right) >\rho_{\bar\theta}^{\mathbf{H}}\left(C_{T} |{\tilde\sigma^{\mathbf{H}},\sigma^{\mathbf{*M}}, \pi_1, h_1}\right).$$ Then the monotonicity property implies that the coordinator's objective function can be reduced using the strategy $g_{\bar\theta}^{\mathbf{H}} = \tilde\sigma^{\mathbf{H}}$. Thus we arrive at the desired contradiction that the solution to the coordinator problem is optimal.

\section{Numerical Study} \label{sec:numbericaldetails}
The figure below depicts optimal paths for three values of the risk aversion parameters: from left to right, $\theta=.01$, $\theta = .1$ and $\theta = .2$. Each edge $(i,j)$ is labeled with the mean and variance of its cost $(\mu_{ij}, \sigma_{ij}^{2})$ and the terminal nodes are labeled with reward distributions $R = (\mu_{i}, \sigma_{i}^{2})$. The three terminal nodes labeled with reward distributions are the only nodes where the human-machine tandem can choose the ``STOP'' action.

\begin{figure}[ht!]
\begin{center}
\resizebox{1.75in}{!}{\includegraphics{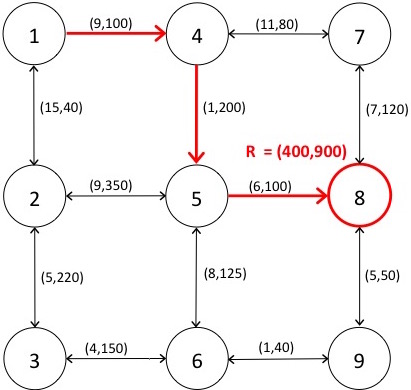}}
\resizebox{1.75in}{!}{\includegraphics{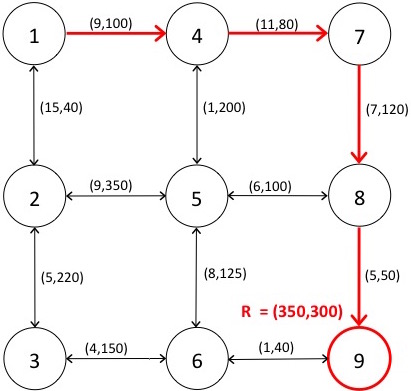}}
\resizebox{1.75in}{!}{\includegraphics{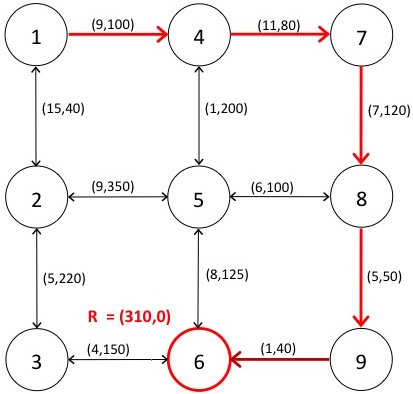}}
\end{center}
\end{figure}

We mathematically define the regret for each of the strategies discussed in Section 5.1. First, we define the best-case policy as the one in which the machine maximizes the correct human's criterion, that is:
$$
BCP = \sum_{i=1}^K \pi_1(\theta_k) \max_{\left\{a_\tau^H, a_\tau^M\right \}} \left(\Ex \left[\sum_{\tau=1}^T c(s_{\tau},a_{\tau}^{H},a_{\tau}^{M}) | \theta=\theta_k \right] + \theta_k Var \left[\sum_{\tau=1}^T c(s_{\tau},a_{\tau}^{H},a_{\tau}^{M}) | \theta=\theta_k \right] \right),
$$
where $K$ is the number of distinct admissible values of the parameter $\theta$. The regret of the ``Machine-Neutral'' strategy is defined by
$$
\Ex^{\pi_1} \left[ \Ex \left[\sum_{\tau=1}^T c(s_{\tau},a_{\tau}^{*H},a_{\tau}^{*M,0} ) \right] + \theta Var \left[\sum_{\tau=1}^T c(s_{\tau}, a_{\tau}^{*H},a_{\tau}^{*M,0}) \right] \right] - BCP,
$$
where the first expectation above is computed under the policy in which the machine incorrectly maximizes the expected cumulative cost, i.e. it assumes $\theta=0$. We denote by
$a_{\tau}^{*M,0}$ the corresponding optimal action chosen by the machine. The regret of the ``Machine-Average'' strategy is defined by
$$
\Ex^{\pi_1} \left[ \Ex \left[\sum_{\tau=1}^T c(s_{\tau},a_{\tau}^{*M,n} ) \right] + \theta Var \left[\sum_{\tau=1}^T c(s_{\tau}, a_{\tau}^{*M,n} ) \right] \right] - BCP,
$$
where the first expectation above is computed under the policy in which only the machine takes actions and the human never intervenes. We denote by $a_{\tau}^{*M,n}$ the corresponding optimal action by the machine. The regret of the ``Human-Machine'' strategy evaluates the performance of our model against the best-case policy. The regret is defined as
$$
\Ex^{\pi_1} \left[ \Ex \left[\sum_{\tau=1}^T c(s_{\tau},a_{\tau}^{*H}, a_{\tau}^{*M}) \right] + \theta Var \left[\sum_{\tau=1}^T c(s_{\tau}, a_{\tau}^{*H}, a_{\tau}^{*M}) ) \right] \right] - BCP,
$$
where the first expectation above is computed under the joint optimal human-machine policy, and $a_{\tau}^{*H}, a_{\tau}^{*M}$ denote the corresponding optimal actions to the risk-POMDP.

For each policy, the optimal value of the coordinator risk-POMDP is solved by expressing the belief profile of the machine as a belief-state and solving for the optimal value function via backwards induction.

\subsubsection*{Acknowledgments}
This research was developed with funding from the Defense Advanced Research Projects Agency (DARPA).

\bibliographystyle{apalike}
\bibliography{Bibliography}
\end{document}